\documentclass{article}
\usepackage{spconf,amsmath,graphicx}
\usepackage{times}
\usepackage{epsfig}
\usepackage{graphicx}
\usepackage{amsmath}
\usepackage{amssymb}
\usepackage{color}
\usepackage{multirow}
\usepackage{dsfont}

\newcommand{\ie}{\textit{i.e.~}}
\newcommand{\myparagraph}[1]{\vspace{5pt}\noindent\textbf{#1}}

\title{
Best sources forward:\\ domain generalization through source-specific nets}
\name{Massimiliano Mancini$^{1,2}$ \qquad Samuel Rota Bul\`o$^{3}$ \qquad Barbara Caputo$^{1,4}$ \qquad Elisa Ricci$^{2,5}$}
\address{$^{1}$Sapienza University of Rome, Rome, Italy\\$^{2}$Fondazione Bruno Kessler, Trento, Italy\\$^{3}$Mapillary Research, Graz, Austria\\$^{4}$Italian Institute of Technology, Genova, Italy\\$^{5}$University of Trento, Trento, Italy
\thanks{This work was supported by the ERC grant 637076 - RoboExNovo and project \textit{DIGIMAP}, grant 860375, funded by the Austrian Research Promotion Agency (FFG).}}

\begin{document}
\begin{titlepage}
\null
\vfill
\renewcommand{\fboxsep}{10pt}
\fbox{\Large\begin{minipage}{\columnwidth}
\textbf{Disclaimer:}

This work has been accepted for publication in the IEEE International Conference on Image Processing:\vspace{4pt}
\newline
link:    https://2018.ieeeicip.org/
\newline
\newline
\textbf{Copyright:} 
\newline
\copyright~2018 IEEE. Personal use of this material is permitted. Permission from IEEE must be obtained for all other uses,  in  any  current  or  future  media,  including  reprinting/  republishing  this  material  for  advertising  or promotional purposes, creating new collective works, for resale or redistribution to servers or lists, or reuse of any copyrighted component of this work in other works.
\newline
\end{minipage}}
\vfill
\clearpage
\end{titlepage}
%
\maketitle
\begin{abstract}
A long standing problem in visual object categorization is the ability of algorithms to generalize across different testing conditions. The problem has been formalized as a covariate shift among the probability distributions generating the training data (\textit{source}) and the test data (\textit{target}) and several domain adaptation methods have been proposed to address this issue. While these approaches have considered the single source-single target scenario, it is plausible to have multiple sources and require adaptation to any possible target domain. This last scenario, named Domain Generalization (DG), is the focus of our work. Differently from previous DG methods which learn domain invariant representations from source data, we design a deep network with multiple domain-specific classifiers, each associated to a source domain. At test time we estimate the probabilities that a target sample belongs to each source domain and exploit them to optimally fuse the classifiers predictions. To further improve the generalization ability of our model, we also introduced a domain agnostic component supporting the final classifier. 
Experiments on two public benchmarks demonstrate the power of our approach. 
\end{abstract}
\vspace{-0.2cm}
\section{Introduction}
\label{sec:intro}
\vspace{-0.4cm}
From self-driving cars to assistive technologies for the cognitive and physically impaired, today we witness a pressing demand for visual recognition systems able to cope with the challenges of unconstrained settings. A crucial component for such systems is their ability to generalize across visual domains, \textit{i.e.} to be able to achieve strong performances regardless of the underlying statistic of the data used for training (source domains), compared to the statistic of all the possible future test data (target domains). While the computer vision community has been aware for quite some time of the existence of a dataset bias issue when considering different data collections \cite{khosla2012undoing}, most efforts have been focused on reducing the domain shift among \textit{two} distributions, corresponding to a \textit{specific source} and a \textit{specific target} domain. Such research efforts go under the name of Domain Adaptation, for which there is a large literature of shallow and deep approaches \cite{csurka2017domain,carlucci2017autodial,xu2018deep,mancini2018boosting}.

A less researched direction is {Domain Generalization (DG)}, that consists in bridging the domain gap regardless of the target data distribution. This concretely corresponds to scenarios where it is costly to acquire in advance target data or it is impossible to predict a priori the specific target scenario where the system will operate. Besides practical considerations, DG attempts to address the issue of dataset bias in a more principled manner: a visual recognizer ready to work in the wild should guarantee robust performances \textit{in any target domain}. 

\begin{figure}[t]
\centering
\includegraphics[width=0.95\columnwidth]{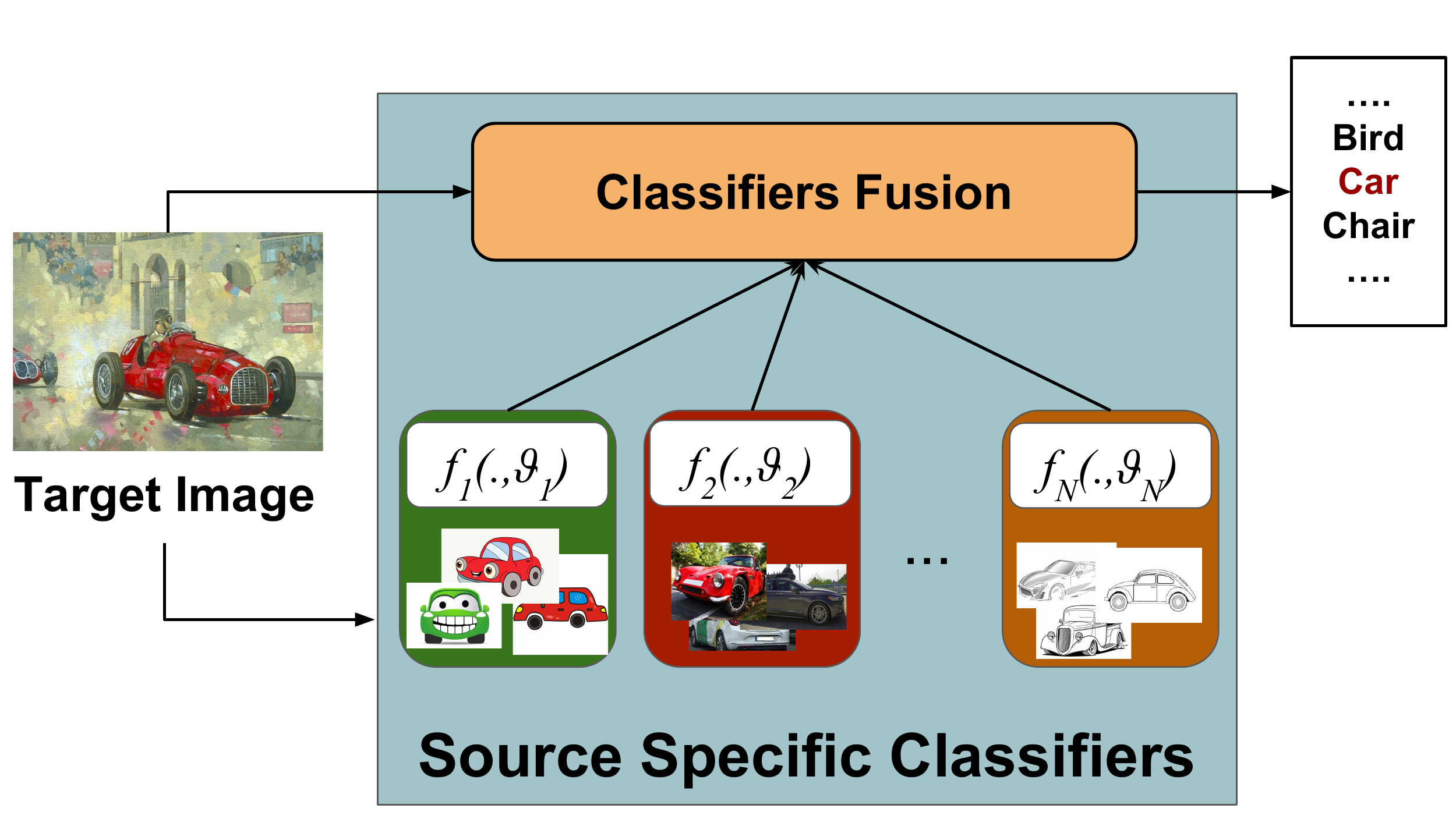}
\vspace{-0.3cm}
    \caption{Intuition behind the the proposed framework. Different domain-specific classifiers and the classifiers fusion are learned at training time on source domains, \textit{in a single end-to-end trainable architecture}. When a target image is processed, our deep model optimally combines the source models in order to compute the final prediction.
} 
   \label{fig:idea}
   \vspace{-0.6cm}
\end{figure}

This paper contributes to this last research thread and we propose a novel deep network for addressing the problem of DG. Different from previous works for DG based on learning domain-invariant representations with deep architectures \cite{li2017deeper,ghifary2015domain}, our intuition is that, given several source domains and their associated domain-specific classifiers, generality can be achieved at test time classifying each incoming target image by optimally fusing the prediction scores of the source-specific classifiers. 
This is achieved through an end-to-end trainable deep architecture with two main components (Fig.\ref{fig:idea}). The first implements the source-specific classifiers, while the second module is a network branch which computes the similarities of an input sample to all source domains, such as to assign weights to the source classifiers and properly merge their predictions. The second module is also designed in order to easily permit, if needed, the integration of a domain agnostic classifier which, acting in synergy with the domain-specific models, can further improve generalization. 
%
We assess our method on two public available datasets, obtaining state of the art performances. 




\myparagraph{Related Work.} 
Although less researched than domain adaptation, the need for DG algorithms has been recognized for quite some time in the literature \cite{muandet2013domain}. 
The works presented so far can be roughly divided in two categories. The first category is based on the intuition that DG can be achieved by abstracting from the available sources some knowledge about the classes to be recognized that is domain independent. This idea is exploited by previous
methods searching for a domain invariant feature space where to project the data \cite{ghifary2015domain,muandet2013domain} or 
by approaches attempting to generate domain agnostic classifiers using both shallow \cite{tommasi2012accv,grubinger2017multi} or deep learning models \cite{li2017deeper,motiian2017unified,li2017domain}.

The second category exploits the idea that it is possible to measure the similarity among the available source domains and every sample of the target domain. 
By exploiting this information robust classification models for the target domain are built constructing different source-specific classifiers and optimally combining them \cite{xu2014exploiting,mancini2018robust}.
Our work falls into this second category. However, opposite to previous studies, we cast the idea into a deep learning framework, proposing to our knowledge one of the first end-to-end trainable deep architecture for DG maintaining source-specific representations. \vspace{-0.4cm} 

\section{Domain Generalization with Source-Specific Classifiers}

\begin{figure}[t]
\centering
\includegraphics[width=0.84\columnwidth]{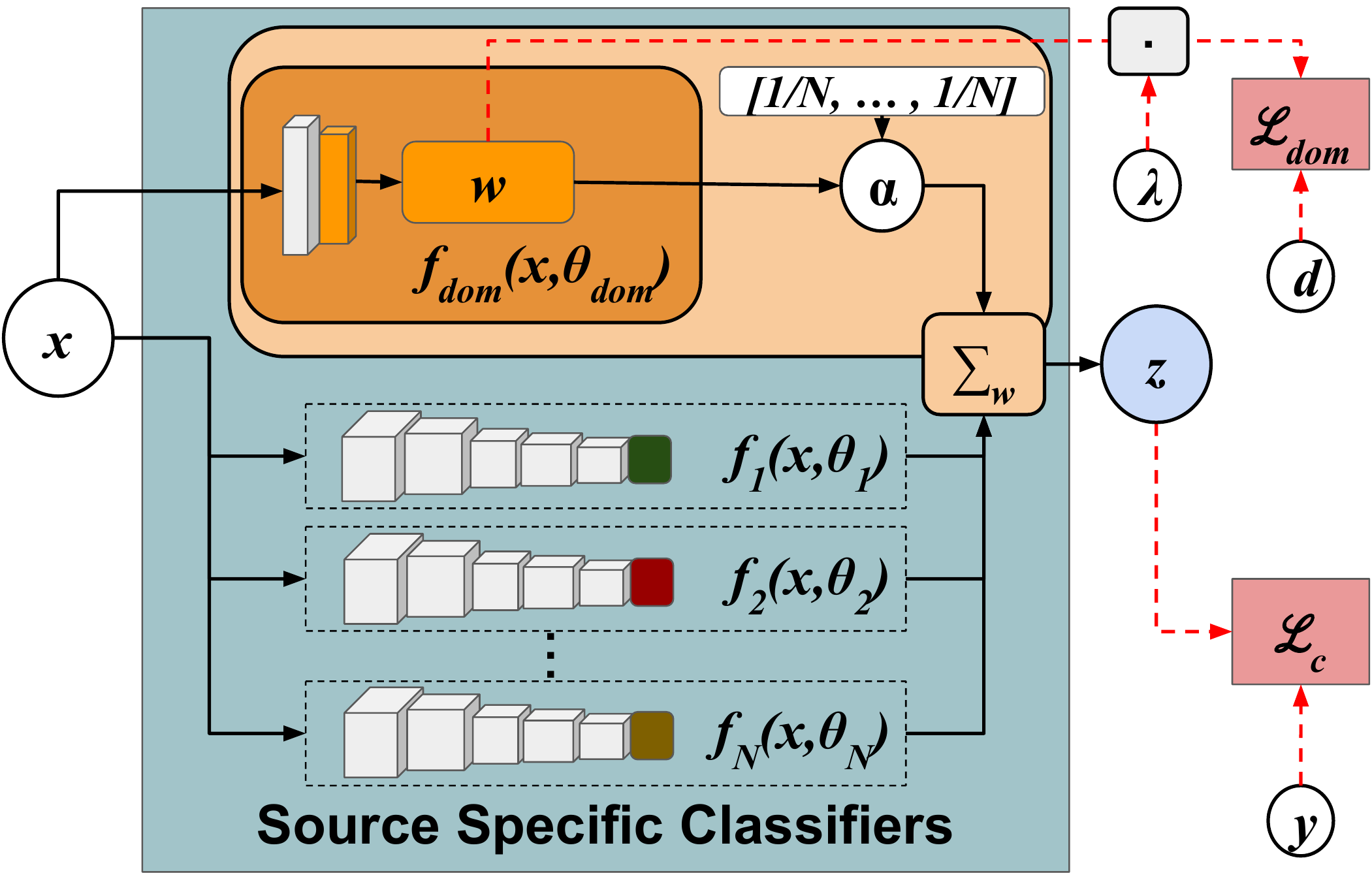}
\vspace{-0.3cm}
    \caption{Simplified architecture of the proposed framework. The input image is fed to a series of domain specific classifiers and to the domain prediction branch. The latter produces the assignment $w$ which is fed to the domain prediction loss. The same $w$ is modulated by $\alpha$ before being used to combine the output of each classifier. The final output of the architecture $z$, is fed to the classification loss. 
\vspace{-0.4cm}    } 
   \label{fig:method}
\end{figure}
\vspace{-0.2cm}
\subsection{Problem Definition and Notation} \vspace{-0.3cm}
The goal of DG is to extend the knowledge acquired from a set of source domains to any unknown target domain. Specifically, let us consider a classification task where $C$ is the number of categories. Our purpose is to learn a classification function $f$ for the unknown target domain, having only access to source samples at training time. Formally, we denote with $\mathcal{X}_T$ the target domain and with $\mathcal{X}_S=\{\mathcal{X}^1_S,\dots,\mathcal{X}^N_S\}=\{(x_i, y_i, d_i)\}_{i=1}^M$  the set of $N$ source domains. Here, $x_i$ is an image, 
$y_i \in \Re^C$ a binary vector with a single non-zero entry indicating the semantic category associated to $x_i$, $d_i \in \{1,\dots, N\}$ the label denoting the domain to which the sample belongs, $M$ the total number of source samples. 

\vspace{-0.3cm}
\subsection{Combining Source-specific Classifiers for DG}
\vspace{-0.3cm}
As discussed in the introduction, 
our aim is to address the DG problem classifying each sample of the target domain through an appropriate combination of domain specific models derived from source domains. Formally, we consider a classification function $f(x_i, \Theta)$ mapping an input image $x_i$ to a vector 
$z_i \in \Re^C$
of class predictions, \ie: 
\vspace{-0.3cm}
\begin{equation}
\label{eq:classifier}
z_i = f(x_i,\Theta)= \sum_{j=1}^N w_{i,j} f_j(x_i,\theta_j)
\vspace{-0.25cm}
 \end{equation}
 where $f_j$ indicates the classifier associated to the $j$-th domain, $\Theta=\{\theta_j\}_{j=1}^N$ denotes the set of parameters to learn and each $\theta_j$ the parameters corresponding to a specific domain $j$. While several choices of $f$ are possible, we choose to implement it with a convolutional neural network with many parallel branches, each corresponding to a domain-specific classifier $f_j$. To avoid high computational cost, we consider the same architecture for each $f_j$, sharing all the parameters 
except those of the last layer before the classifier. In formulas $\theta_j=\{\hat{\theta}_s,\hat{\theta}_j\}$, where $\hat{\theta}_s$ indicates the shared parameters and $\hat{\theta}_j$ the domain-specific ones.

The computation of the weights $w_{i,j}$ used to combine the domain-specific classifiers is at the core of our method and is discussed in the following.

In case of source samples, since each image is associated to a specific domain there is an obvious choice for setting the weights $w_{i,j}$. Specifically, we can define $w_{i,j}=\mathds{1}_{d_i=j}$
, where $\mathds{1}_{d_i}$ is an indicator function with value 1 if $d_i=j$ and 0 otherwise. 
This corresponds to learning $N$ domain-specific classifiers independently. In fact, due to the presence of the indicator function, training the network using a standard classification loss $\mathcal{L}_{c}(z_i,y_i)$, \textit{e.g.} a cross-entropy loss, 
we have $\frac{\partial \mathcal{L}_{c}}{\partial f_j(x_i,\theta_j)}=0$ for $j\neq d_i$. As a consequence, the parameters $\hat\theta_j$ of the classifier relative to domain $j$ are updated using only the losses computed on samples belonging to the set $\mathcal{X}_{S}^J$. 

Unfortunately, learning domain-specific classifiers as discussed above is not useful in a DG scenario. In fact, while at training time we can rely on the domain label $d_i$ to build the indicator functions, this is not possible at test time since we do not have this information for target samples. To solve this issue and learn the weights $w_{i,j}$, we propose to incorporate into our deep architecture a parallel network branch, \textit{i.e.} a \textit{domain prediction} branch, mapping a given input image $x_i$ to the associated weight vector $w_i=[w_{i,1},\cdots,w_{i,N}]^T\in \Re^N$. In other words we define $w_i=f_{dom}(x_i,\theta_{dom})$,
where $f_{dom}$ is the mapping function corresponding to the domain prediction branch. For each input image $x_i$, we impose $0\leq w_{i,j} \leq 1$ and $\sum_{j=1}^N w_{i,j}=1$ for all $1\leq i\leq N$. Thus, each weight $w_{i,j}$ represents the probability that $x_i$ belongs to domain $j$. 

Since at training time we have access to the domain labels $d_i$ of source samples $x_i$, we can learn the parameters $\theta_{dom}$ by minimizing a loss function $\mathcal{L}_{dom}(w_i,\hat{d}_i)$ between $w_i$ and $\hat{d}_i$, where $\hat{d}_i \in \Re^N$ is a binary vector with a single non-zero entry corresponding to the domain label $d_i$.
In our implementation, since $f_{dom}$ shares parameters with the classification branches $f_{j}$, we train the proposed architecture minimizing the following loss function:
\vspace{-0.1cm}
 \begin{equation}
\label{eq:full-loss}
\mathcal{L}=\frac{1}{M}\sum_{i=1}^{M} \left( \mathcal{L}_{c}(z_i,y_i)+\lambda\,\, \mathcal{L}_{dom}(w_i,d_i) \right)
 \end{equation}
where both the semantic classification loss $\mathcal{L}_c$ and the domain prediction loss $\mathcal{L}_{dom}$ are implemented with cross-entropy loss. The hyperparameter $\lambda$ balances the contribution of the semantic classification and the domain prediction terms.



One possible issue with the proposed deep architecture is that, as minimizing the domain loss promotes the learning of independent source classifiers, source sets with few samples may correspond to classifiers with poor performances. While parameter sharing among the deep models implementing $f_j$ naturally limits this effect, we further improve the robustness of our model adding a domain-agnostic component into the final classification function. In practice, we introduce a parameter $0<\alpha<1$ and at training time we randomly switch with probability $\alpha$ between using the computed weights $w_{i,j}$ or assigning to all of them the same value $1/N$. This choice corresponds to modifying the classification model in Eqn.(\ref{eq:classifier}) as follows:
\vspace{-0.1cm}
 \begin{equation}
 \label{eq:final-classifier}
 z_i=(1-\alpha) \sum_{j=1}^{N} w_{i,j} f_{j}(x_i,\theta_j) + \frac{\alpha}{N}\sum_{j=1}^{N} f_{j}(x_i,\theta_j)
 \end{equation}
As shown in the formula the parameter $\alpha$ is used to regulate the trade-off between the domain specific and the domain agnostic component. Figure \ref{fig:method} provides an overview of the proposed end-to-end trainable deep architecture.

\section{Experiments}
\vspace{-0.3cm}
\myparagraph{Datasets.}
We test the performance of our method on two publicly available benchmarks. The \textbf{rotated-MNIST} \cite{ghifary2015domain} is a dataset composed by different domains originated applying different degrees of rotations to images of the original MNIST digits dataset \cite{lecun1998gradient}. We follow the experimental protocol of \cite{motiian2017unified}, randomly extracting 1000 images per class from the dataset and rotating them respectively of 0, 15, 30, 45, 60 and 75 degrees counterclockwise. As previous works, we consider one domain as target and the rest as sources. 

The \textbf{PACS} database \cite{li2017deeper} is a recently proposed benchmark which is interesting due to the high domain shift within its domains. It contains images taken from different representations (\ie Photo, Art paintings, Cartoon and Sketches) associated to seven semantic categories. Following the experimental protocol of \cite{li2017deeper}, we train our model considering three domains as source datasets and the remaining one as target. 
\vspace{-0.2em}

\myparagraph{Networks and training protocols.}
In our evaluation we set the parameters $\alpha=0.25$ and $\lambda=0.5$. 
For the experiments on the rotated-MNIST dataset, we employ the LeNet architecture \cite{lecun1998gradient} following \cite{motiian2017unified}. The network is trained from scratch, using a batch size of 250 with an equal number of samples for each source domain.  We train the network for 10000 iterations, using Stochastic Gradient Descent (SGD) with an initial learning rate of 0.01, momentum 0.9 and weight decay 0.0005. The learning rate is decayed through an inverse schedule, following previous works \cite{ganin2015unsupervised}. For the domain prediction branch, we take as input the image and perform two convolutions, with the same parameters of the first two convolutional layers of the main network. Each convolution is followed by a ReLU non linearity and a pooling operation. 
The domain prediction branch terminates with a global average pooling followed by a fully connected layer which outputs the final weights. To ensure that $\sum_{j=1}^N w_{i,j}=1$, we apply the softmax operator after the fully connected layer. 
 
For PACS, we trained the standard AlexNet architecture, starting from the ImageNet pretrained model. 
We use a batch size of 192, with 64 samples for each source domain. The initial learning rate is set to $5\cdot10^{-4}$ with a weight decay of $10^{-6}$ and a momentum of 0.9. We train the network for 3000 iterations, decaying the initial learning rate by a factor of 10 after 2500 iterations, using SGD. For the domain prediction branch, we use the features of \texttt{pool5} as input, performing a global average pooling followed by a fully-connected layer and a softmax operator which outputs the domain weights.

\begin{figure}[t]
\centering
\includegraphics[width=0.8\columnwidth,trim=0.0cm 2cm 0cm 0cm,]{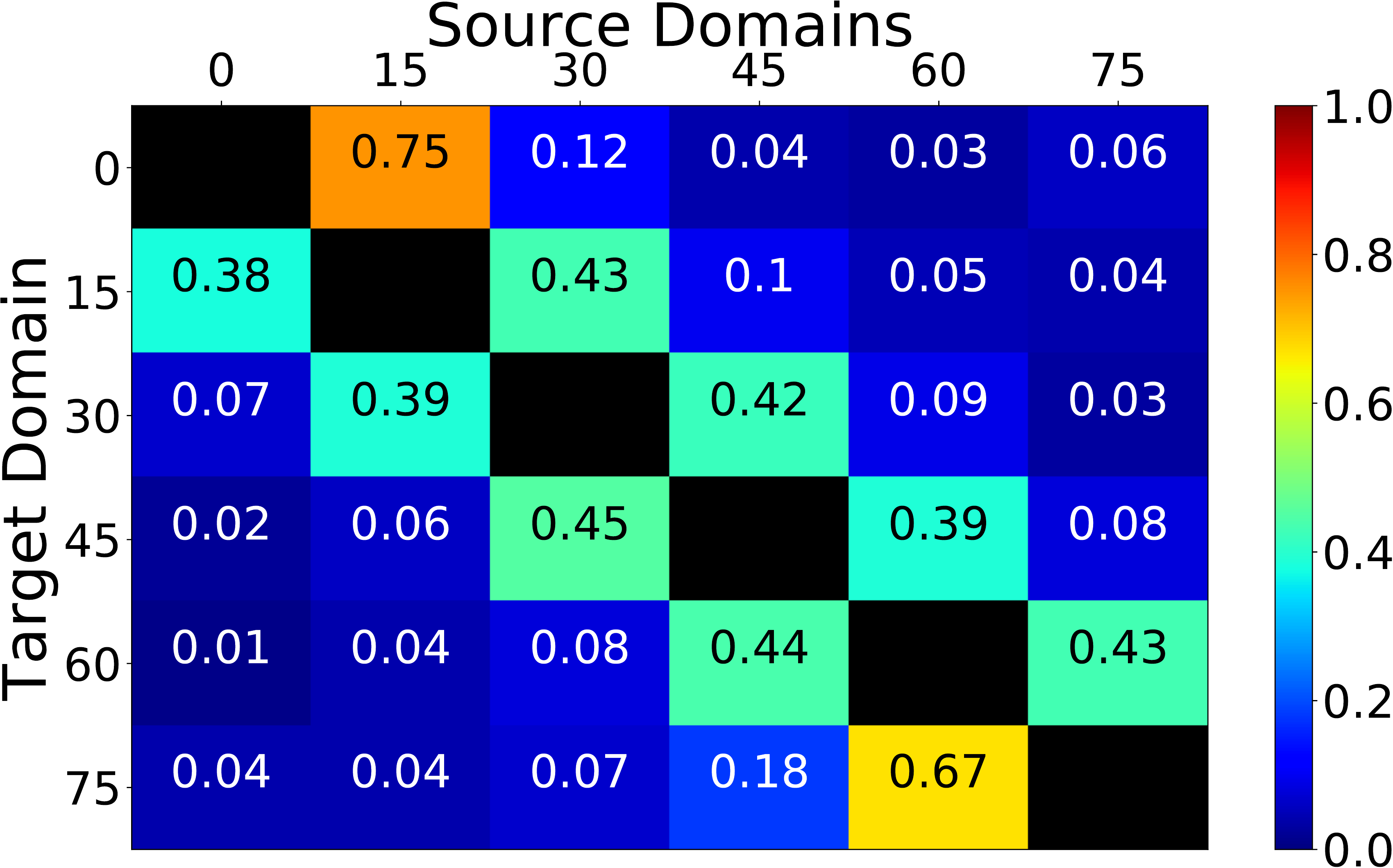}
    \caption{Rotated-MNIST dataset: analysis of the assignments computed by the domain prediction branch.} 
   \label{fig:cm-mnist}
   \vspace{-0.45cm}
\end{figure}

Our evaluation is performed using a NVIDIA GeForce 1070 GTX GPU, implementing all the models with the popular Caffe \cite{jia2014caffe} framework. For the baseline AlexNet architecture we take the pretrained model available in Caffe.
\vspace{-0.2em}
\myparagraph{Results.} 
We first test the effectiveness of our model on the rotated-MNIST benchmark. 
We compare our approach with the method in \cite{motiian2017unified} and the multi-task autoencoders in \cite{ghifary2015domain} and \cite{rifaiexplicit}. The results from baseline methods are taken directly from \cite{motiian2017unified}.
\begin{table}[t]
			\caption{Rotated-MNIST dataset: comparison with previous methods. } 
		\centering
		\scalebox{.9}{
		\begin{tabular}{ l | c  c  c  c  c  c | c } 
			\hline
			Method & 0 & 15 & 30 & 45 & 60 & 75& Mean\\
            \hline
             \cite{rifaiexplicit} & 72.1 & 95.3 & 92.6 & 81.5& 92.7&79.3&85.5 \\ 
             \cite{ghifary2015domain} & 82.5 & \textbf{96.3} & 93.4 & 78.6& 94.2&80.5&87.5 \\ 
             \cite{motiian2017unified} & 84.6 & 95.6 & 94.6 & 82.9& 94.8&82.1&89.1 \\ 
             Ours & \textbf{85.6}& 95.0 & \textbf{95.6}& \textbf{95.5}& \textbf{95.9}&\textbf{84.3}&\textbf{92.0} \\ \hline
             
		\end{tabular}
        }
		\label{tab:mnist}
        \vspace{-0.25cm}
\end{table}
As shown in Table \ref{tab:mnist}, our model outperforms all the baselines. A remarkable gain in accuracy is achieved in the $45^o$ case. 
We ascribe this gain to the capability of our deep network to assign, for each target image, more importance to the source domains corresponding to the closest orientations, increasing the weights of the associated classifiers. Indeed, since $45^o$ is in the middle of the range between all possible orientations, it is likely that a stronger classifier can be constructed since we can exploit all the source models appropriately re-weighted. To further verify the effectiveness of our framework and its ability to properly combine source-specific models, we also compute for target samples with different orientations the number of assignments to each source domain. In this experiment one target sample $x_i$ is assigned to a source domain by computing the $\arg \max_j w_{i,j}$. The results are shown in Fig. \ref{fig:cm-mnist} (the number of assignments are normalized for each row). The figure clearly shows that the proposed domain prediction branch tends to associate a target sample to the source domains corresponding to the closest orientations. Consequently, our deep network classifies target samples constructing a model from the most related source classifiers. This results into more accurate predictions than previous domain-agnostic models due to the specialization of source classifiers on specific orientations. 

We also perform experiments on the PACS dataset. We compare our model with both previous approaches using precomputed features (in this case DECAF-6 features \cite{donahue2014decaf}) as input \cite{ghifary2015domain,xu2014exploiting,muandet2013domain} and end-to-end trainable deep models \cite{li2017deeper,li2017learning}. For a fair comparison the deep models \cite{li2017deeper,li2017learning} and our network are all based on the same architecture, \ie AlexNet.  Table \ref{tab:pacs} shows the results of our comparison. The performance of previous methods are taken directly from previous papers \cite{li2017deeper,li2017learning}. For our approach and \cite{li2017deeper} we also report results obtained without finetuning.
Our model outperforms all previous methods. 
These results are remarkable because, differently from the rotated-MNIST dataset, in PACS the domain shift is significant and it is not originated by simple image perturbations. Therefore, the association between a target sample and the given source domains is more subtle to capture. For sake of completeness we also report the performances obtained with the standard AlexNet network. These results shows that state of the art deep models have excellent generalization abilities, typically outperforming shallow models. However, designing deep networks specifically addressing the DG problem as we do leads to higher accuracy.

\begin{table}[t]
			\caption{PACS dataset: comparison with previous methods.} 
		\centering
		\scalebox{.9}{
		\begin{tabular}{ l | c c c c | c } 
			\hline
			Model & Art & Cartoon & Photo & Sketch & Mean\\\hline
             \cite{ghifary2015domain}& 60.3 & 58.7 & 91.1 & 47.9 & 64.5  \\ 
             \cite{xu2014exploiting}& 59.7 & 52.9 & 85.5 & 37.9 & 58.9  \\  
               \cite{muandet2013domain}& 64.6 & 64.5 & 91.8 & 51.1 & 68.0  \\ 
               \hline
            \cite{li2017deeper} (no ft) & 62.7 & 52.7 & 88.8 & 52.2& 64.1 \\ 
       \cite{li2017deeper}& 62.9&\textbf{67.0}&89.5&57.5&69.2\\
            \cite{li2017learning}&\textbf{66.2}&66.9&88.0&59.0&70.0\\   
        Ours (no ft) & 64.1 & 60.6 & \textbf{90.4} & 49.4 & {66.1} \\
          Ours &64.1& 66.8 & 90.2&\textbf{60.1} & \textbf{70.3}  \\ \hline
         AlexNet  \cite{li2017deeper}   &63.3&63.1&87.7&54.1&67.1\\ \hline

		\end{tabular}
        }
		\label{tab:pacs}
        \vspace{-0.5cm}
\end{table}


We also perform a sensitivity analysis to study the impact of the parameter $\alpha$ on the performance and demonstrate the benefit of adding a domain-agnostic classifier. 
We consider the proposed approach without finetuning.
\begin{table}[t]
			\caption{PACS dataset: sensitivity analysis. 
            } 
		\centering
		\scalebox{.9}{
		\begin{tabular}{ l | c  c  c  c  } 
			\hline
			$\alpha$ & Art & Cartoon & Photo & Sketch \\\hline
             0 & \textbf{65.2} & 54.5 & \textbf{90.7} & \textbf{52.4} \\ 
 0.25 & 64.1 & 60.6 & 90.4 & 49.4 \\ 
 0.5 & 63.8 & \textbf{61.0} & 90.4 & 49.1 \\ 
             0.75 & 64.0 & 60.9 & 90.5 & 47.8\\
             1 & 63.0 & 60.1 & 90.5 & 47.5 \\ \hline
		\end{tabular}
        }
		\label{tab:ablation-alpha}
    \vspace{-0.5cm}
\end{table}
As shown in Table \ref{tab:ablation-alpha}, considering only the source-specific classifiers ($\alpha=0$) leads, on average, to the best performances, surpassing in the majority of the cases a domain agnostic classifier obtained by setting $\alpha=1$. This confirms our original intuition that addressing DG by fusing multiple source models is an effective strategy. However, there are few situations where using only source models can lead to a decrease in accuracy (\textit{e.g.} in the setting Cartoon) and incorporating a domain-agnostic component, even with reduced weight as $\alpha=0.25$, improves generalization accuracy.  
\vspace{-0.3cm}
\section{Conclusions}
\vspace{-0.3cm}
We presented a novel deep architecture for addressing the problem of DG by exploiting multiple domain-specific classifiers. In the network a domain prediction branch chooses the optimal combination of source classifiers to use at test time, based on the similarity between the input image and the samples from the source domains. A domain agnostic component is also introduced in our framework further improving the performance of our method. Our experiments demonstrate the effectiveness of the proposed deep architecture which outperforms state of the art models in two benchmarks. Future works will include 
the exploration of different architectural choices for the domain prediction branch.

\bibliographystyle{ieee}
\bibliography{root}

\end{document}